%% file: Main.tex
\documentclass[10pt,journal,compsoc]{IEEEtran}
\ifCLASSOPTIONcompsoc
  \usepackage[nocompress]{cite}
\else
  \usepackage{cite}
\fi

\ifCLASSINFOpdf
\else
\fi
\usepackage{algorithm} %http://ctan.org/pkg/algorithms
\usepackage{algpseudocode}% http://ctan.org/pkg/algorithmicx
\algnewcommand{\Initialize}[1]{%
  \State \textbf{Initialize:}
  \Statex \hspace*{\algorithmicindent}\parbox[t]{.8\linewidth}{\raggedright #1}
}
\usepackage{multirow}
\usepackage{tabularx,booktabs}
\usepackage{amsmath,amsfonts,amssymb}
\usepackage{subfig}
\usepackage{adjustbox}
\usepackage{xcolor}
\usepackage{comment}

\AtBeginDocument{%
\providecommand\BibTeX{{%
    \normalfont B\kern-0.5em{\scshape i\kern-0.25em b}\kern-0.8em\TeX}}}
\newtheorem{definition}{Definition}  
\usepackage{url}
\usepackage{lipsum}

\begin{document}

\title{Deep Graph Memory Networks for Forgetting-Robust Knowledge Tracing}

\author{Ghodai Abdelrahman,
        Qing Wang
\IEEEcompsocitemizethanks{\IEEEcompsocthanksitem All authors are with Research School of Computer Science, 
 Australian National University, Canberra,
ACT, 0200.\protect\\

}% <-this % stops an unwanted space
}


\markboth{Journal Submission }%
{Shell \MakeLowercase{\textit{et al.}}: Bare Demo of IEEEtran.cls for Computer Society Journals}

\IEEEtitleabstractindextext{%
\begin{abstract}
 Tracing a student's knowledge is vital for tailoring the learning experience. Recent knowledge tracing methods tend to respond to these challenges by modelling knowledge state dynamics across learning concepts. However, they still suffer from several inherent challenges including: modelling forgetting behaviours and identifying relationships among latent concepts. To address these challenges, in this paper, we propose a novel knowledge tracing model, namely \emph{Deep Graph Memory Network} (DGMN). In this model, we incorporate a forget gating mechanism into an attention memory structure in order to capture forgetting behaviours dynamically during the knowledge tracing process. Particularly, this forget gating mechanism is built upon attention forgetting features over latent concepts considering their mutual dependencies. Further, this model has the capability of learning relationships between latent concepts from a dynamic latent concept graph in light of a student's evolving knowledge states. A comprehensive experimental evaluation has been conducted using four well-established benchmark datasets. The results show that DGMN consistently outperforms the state-of-the-art KT models over all the datasets. The effectiveness of modelling forgetting behaviours and learning latent concept graphs has also been analyzed in our experiments.
\end{abstract}

}

\maketitle

\IEEEdisplaynontitleabstractindextext

% \fi
\IEEEpeerreviewmaketitle

\input{Introduction.tex}
\input{Problem}

\input{Methodology}
\input{Experiments}

\input{Results}

\input{RelatedWork}

\input{Conclusion.tex}

%\newpage
\bibliographystyle{IEEEtran}
\bibliography{References}

\end{document}

%% file: Introduction.tex
\section{Introduction}
\label{sec:introduction}

Knowledge tracing (KT) is a fundamental problem in human learning and has attracted increasing interests in recent years. Conceptually, the KT problem is to predict the probability of correctly answering new questions given a history of question answering sequence~\cite{Corbett1994}. 
Early work has attempted to address the KT problem using Bayesian approaches~\cite{Corbett1994,Baker2008,Pardos_2011,Yudelson_2013}. Typically, these approaches used state-space models (e.g., hidden Markov models) to track a knowledge state as a binary random variable (i.e., 0 for not knowing and 1 for knowing) and predict the probability of correctly answering a question. However, one major limitation of these approaches is that they rely on oversimplified assumptions about knowledge states and latent concepts in order to keep the Bayesian inference computationally tractable. 
To overcome this limitation, deep learning KT approaches have been proposed. For example, Piech et al.~\cite{DKT2015} introduced a recurrent neural network called \emph{Deep Knowledge Tracing} (DKT) to track knowledge states over a question answering sequence. Several recent attempts~\cite{DKVMN17,SKVMN,AKT20} further enhanced deep KT models with an external memory structure to represent knowledge states. These deep learning models have shown the state-of-the-art results on well-established KT datasets. Nonetheless, they still suffer from the following inherent challenges:
\begin{itemize}
\item[(1)] Cognitive modelling studies on human's memory showed a knowledge decline trend over time due to forgetting behaviours, known as the \emph{forget curve theory}~\cite{ebbinghaus2013}. However, it is challenging to model such forgetting behaviours (e.g., when and how forgetting behaviors occur) into knowledge states and reflect them in a KT model for predicting a student's learning performance.%, particularly when a sequence of question answering activities is long.
\item[(2)] Concept learning is a phenomenal ability of human beings, which involves to extract latent concepts from questions (e.g., learning topics in a math course) and identify their relationships \cite{novak2008theory}. However, it is not yet clear to which extent latent concepts and their relationships can be explicitly modelled and leveraged for predicting a student's learning performance in a KT model. 
\end{itemize}
%focus primarily on modelling temporal dynamics of the knowledge state across different latent concepts without paying attention to the dependency relationships among them. -- limitations: relationships among latent concepts are not captured, forgetting is not well represented.

%Nevertheless, human cognitive studies on memory and learning retention behaviours~\cite{Cog_forget_06,Cog_forget_11} showed that a student's knowledge declines overtime due to forgetting behaviour, which is known as the \emph{Forget curve theory} \cite{ebbinghaus2013}. %existing methods depend solely on dependencies among questions in a question answering sequence to trace a knowledge state.

In this work, we aim to tackle these challenges by introducing a novel forgetting-robust KT model, namely \emph{Deep Graph Memory Network} (DGMN). 
 The novelty of this proposed model lies in three aspects: (1) modelling  forgetting features over latent concepts in a concept space while considering their mutual relationships; (2) learning a dynamical latent concept graph to reflect the dynamics of a student's knowledge states over latent concepts and their relationships; (3) incorporating a forget gating mechanism to account for the effect of forgetting features on a student's learning performance. Below, we briefly describe these three new aspects of our proposed DGMN model.

%In a nutshell, DGMN ... provides a unified KT method that couples an external memory structure for modelling forgetting behaviours in evolving knowledge states with a graph representation of latent concepts and their relationships. %The proposed model uses attentive memory structures to learn embedding representations of involved latent concepts and dynamics of the knowledge state over them. %In addition, %it considers relationships among the latent concepts by learning a latent concept graph representation based on the change in the knowledge state memory during training time. 

\begin{figure}[t!]
  \centering
  \includegraphics[width=0.95\linewidth]{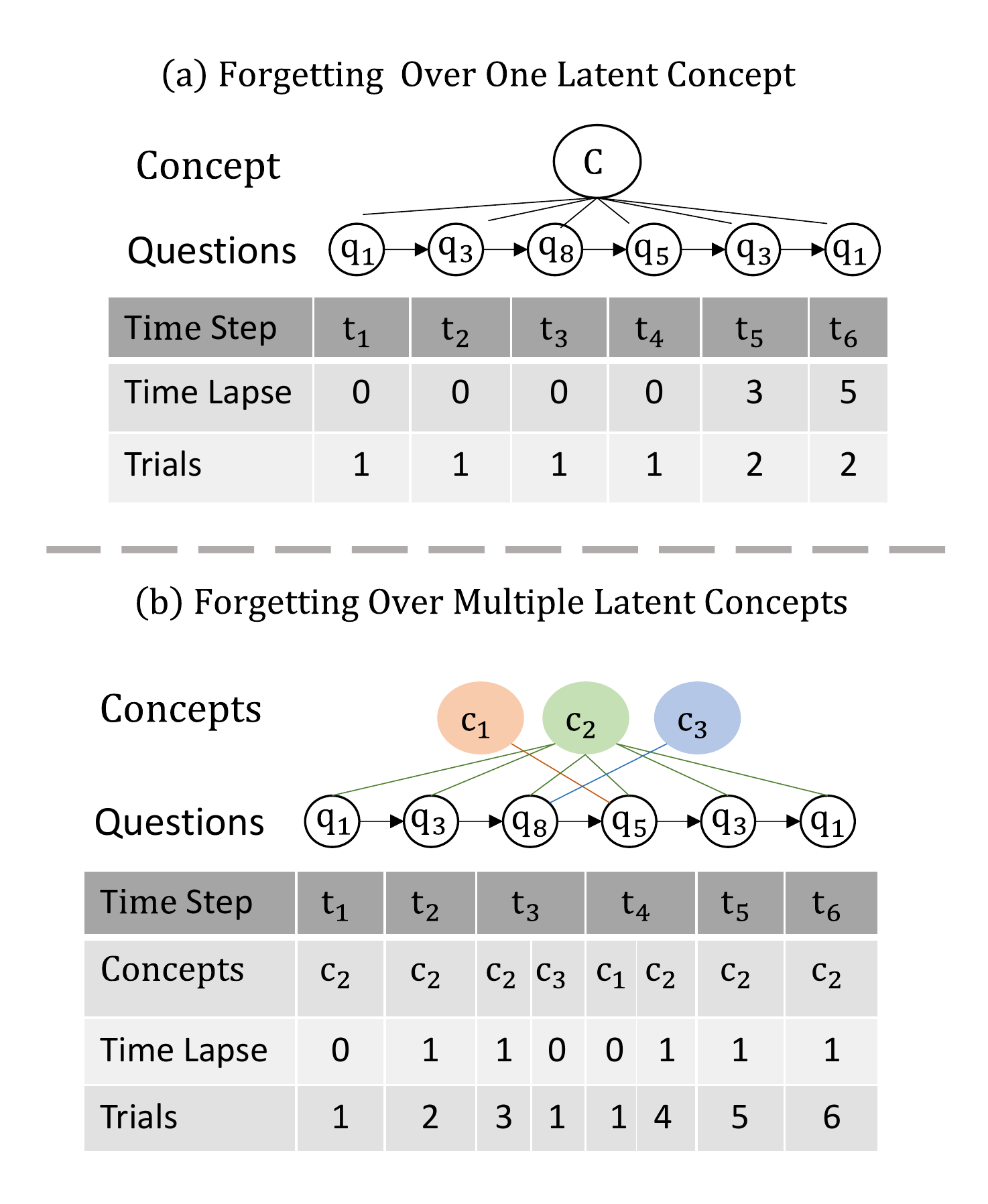}
  \caption{An example for illustrating the difference between modelling forgetting over one latent concept and over multiple latent concepts, where \emph{Time Lapse} and \emph{Trials} are two forgetting features used for modelling forgetting.}% (see a detailed explanation for these forgetting features in Section \ref{sec:introduction}).}
  \label{fig:forgetting}\vspace{-0.3cm}
\end{figure}

Recent KT efforts to address forgetting behaviours tend to have limitations. One direction~\cite{DKT_Forget} follows a deep learning approach in modelling forgetting and it assumes only one latent concept to exist, which is not a realistic assumption as in many real-world cases where would be multiple concepts (e.g., topics such as addition, subtraction, and multiplication in a math course). Moreover, this assumption renders the calculation of forgetting features to be based on questions themselves as all of them share the same concept resulting in an inaccurate estimation for forgetting as shown in the example depicted in Figure~\ref{fig:forgetting}. The other direction~\cite{qforget,forget_cikm17, Zhenya_forget_20} follows a Bayesian approach that assumes a prior information on a student knowledge state across concepts and considers forgetting over concepts but in isolation from their mutual relationships (e.g., prerequisites dependencies between topics in a course) as they lack methods for representing these relationships. The deep learning direction has an advantage over the Bayesian one in bypassing the assumption around existence of prior information about student knowledge states; thus, it can operate in cold-start scenarios. To the best of our knowledge, the work presented in this paper is the first deep learning approach that considers a student forgetting behaviour over multiple concepts and accounts for their mutual relationships.  

In the real world, however, students tend to forget or remember latent concepts (i.e., essential skills) rather than questions themselves and the relationships across these latent concepts would impact the forgetting effect on each of them. For example, a student might be able to correctly answer a new math question which she has never seen before, after she has mastered the latent concepts (e.g., dot product) behind such a question. Figure~\ref{fig:forgetting} illustrates two different forgetting modelling paradigms: \emph{over one latent concept shared across questions} versus \emph{over multiple latent concepts shared across questions}. Let us consider two forgetting features~\cite{DKT_Forget}: (i) \emph{Time Lapse} that counts the number of time steps lapsed since a question has been answered last time,
 and (ii) \emph{Trials} that refers to the total number of times that a student has answered a question. In Figure~\ref{fig:forgetting}.a, forgetting is modelled over one latent concept shared by all the questions that are identified by their question tags, while in Figure~\ref{fig:forgetting}.b, forgetting is modelled over multiple concepts that are identified by their concept tags and shared by the questions. We can see that the question $q_1$ has appeared twice in this sequence. At the time step $t_6$, $q_1$ thus has the forgetting features \emph{Time Lapse}=5 and \emph{Trials}=2 in Figure~\ref{fig:forgetting}.a and \emph{Time Lapse}=1 and \emph{Trials}=6 in Figure~\ref{fig:forgetting}.b. Accordingly, forgetting is considered to likely occur on $q_1$ in Figure~\ref{fig:forgetting}.a. However, since the concept underlying $q_1$ is $c_2$, which indeed underlies all the questions in this sequence (i.e., $q_1$, $q_3$, $q_5$ and $q_8$), the student has learned $c_2$ through all these questions and forgetting would unlikely happen at the time step $t_6$. Therefore, modelling forgetting features over multiple concepts can improve the tracing ability on a student's knowledge states, whereas modelling over one concept cannot achieve this outcome.
 
 Nonetheless, when learning forgetting features over a concept space, a difficulty arises: \emph{since only a question answering sequence is available, how can we model forgetting features effectively in terms of latent concepts behind questions rather than questions themselves?} We observe that, the key to tackling this difficulty is to design a KT model that can not only learn latent concepts behind questions via a question answering sequence, but also capture the relationships among latent concepts. More importantly, such a KT model should have the capability of \emph{dynamically} tracing changes on latent concepts and their relationships in light of a student's performance over time. 
 Previously, several graph-based KT approaches have been suggested~\cite{GIKT20,GKT19}. However, these approaches assume that graphs for latent concepts are predefined based on the ground truth in training data, thereby failing to capture the dynamics of a student's knowledge states in terms of how well they master latent concepts and their connections over time. In this work, our DGMN model overcomes this problem by designing a \emph{deep graph memory structure}, consisting of an attentive key-value memory for dynamically capturing a student's knowledge states over latent concepts and a latent concept graph for dynamically learning relationships among latent concepts.

%These approaches represent latent concepts as nodes and the relationships between latent concepts as edges in a graph. Then, based on such a graph, they predict the probability of correctly answering new questions. However, these approaches lack

%mechanism that can explicitly capture latent concepts and their relationships dynamically during the ....

Even with modelling forgetting features over a concept space upon a deep graph memory structure, there is still one question to be answered: \emph{how can we effectively use forgetting features to reflect a student's forgetting behaviour (i.e., decline of memory retention in time) and count for its effect on predicting answers?} This poses a need to find a good way of combining forgetting features with knowledge state features (i.e., skill mastery information). To address this, our DGMN model incorporates a forget gating mechanism that can learn an optimal combination of these two groups of features during training, thus enabling an integrated analysis of their joint effects on predicating answers. Furthermore, non-linearity introduced by such a forget gating mechanism can enhance the learning capacity of representing informative features in knowledge tracing for predicting a student's learning performance.

%we extract useful information from a latent concept graph to help improve KT predication in two ways: (1) embedding latent concepts and their relationships from a latent concept graph into a graph summary vector for predicating a student's learning performance, and (2) ranking questions in terms of the importance of its underlying latent concepts for improving training effectiveness. 

 %\g{Therefore, modelling forgetting over a concept space enables generalisation over questions sharing relevant concepts. Moreover, it makes the computation cost of forgetting features more efficient as usually the dimensionality of the latent concepts space is less than the one of the question space (i.e., number of questions vs. number of learning skills).}

% It can be also noticed that storing forgetting features over a question space yields higher dimensionality, thus requiring higher computational cost than over a latent concept space.

In summary, the main contributions of this work are as follows:\vspace*{-0.2cm}
\begin{itemize}
    \item We propose a novel KT model, namely DGMN, which is augmented with a deep graph memory structure to dynamically trace both a student's knowledge states over latent concepts in an attentive key-value memory and the relationships among latent concepts in a latent concept graph.
    \item We design the modelling of forgetting features over a latent concept space considering their mutual relationships and devise a forget gating mechanism to combine forgetting features with knowledge state features in our proposed DGMN model for predicting a student's learning performance. %meaningful information
    \item We propose to capture embedding representations of latent concepts and their relationships in a dynamic latent concept graph and leverage its useful information to rank questions for improving model optimization. 
    
    % to accurately predict the probability of correctly answering new questions

    \item We have extensively evaluated the performance of our proposed DGMN model by comparing it against the state-of-the-art deep learning and graph KT models on four well-established KT datasets.
\end{itemize}

The reminder of this paper is organized as follows.  Section \ref{sec:pd} defines the research problem. Section \ref{sec:methodology} describes our proposed KT method. Section \ref{sec:exp} presents the experimental design. Section \ref{sec:res} discusses the experimental results and findings. Section \ref{sec:rl} reviews the related work and Section \ref{sec:conclusion} concludes the paper.

%% file: Problem.tex
\begin{figure*}[t!]
\centering
\includegraphics[scale=0.5]{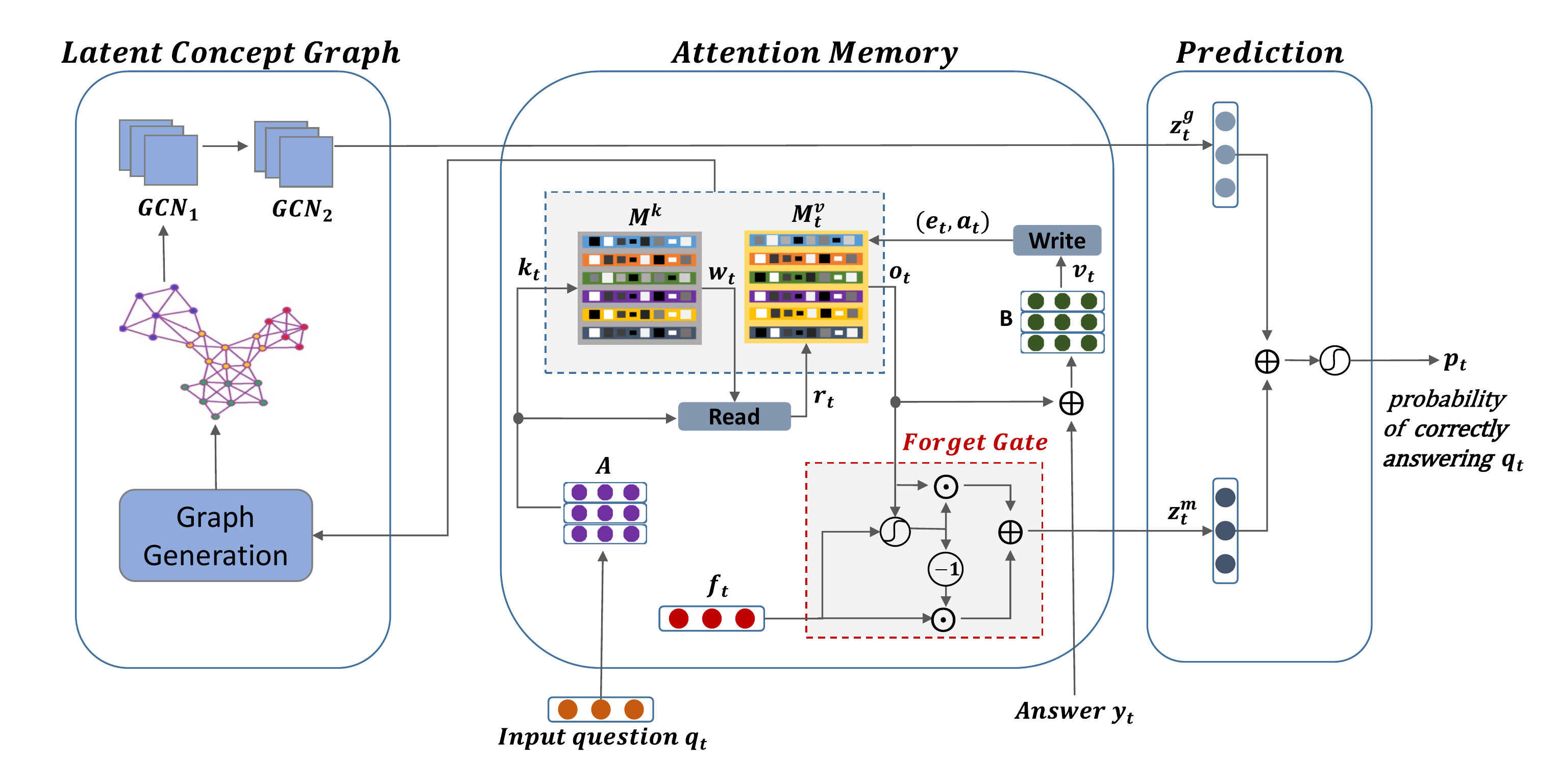}\vspace{-0.3cm}
\caption{Model architecture of our proposed DGMN model, where the latent concept graph (LCG) module is on the left side, the attention memory (AM) module is in the middle, and the answer prediction layer is on the right side.}\vspace{-0cm}
\label{fig:Model}
\end{figure*}

\section{Problem Definition}
\label{sec:pd}

Let $Q$ be a set of questions and $\mathbf{x}_t=(\{q_1,y_1\},\{q_2,y_2\},$ $\dots,\{q_{t-1},$ $y_{t-1}\})$ be a \emph{question answering sequence}, where $q_i\in Q$ and $y_i\in\{0,1\}$ is an answer of $q_i$ with $0$ representing an incorrect answer and $1$ representing a correct answer. A question answering sequence contains the history of a student's question answering interactions. We formulate the KT problem as a sequence learning problem~\cite{DKVMN17}. 

\begin{definition}
 Given a question answering sequence $\mathbf{x}_t$ and an input question $q_t\in Q$, the \emph{knowledge tracing (KT) problem} is to predict the probability that $q_t$ is answered correctly. %, i.e, $p(y_t=1|\mathbf{x}_t,q_t)$. 
\end{definition}

%Each knowledge tracing problem is associated with a set of questions $Q$, from which a question $q_t$ is taken. 
In the KT problem, the questions in $Q$ correspond to a number of \emph{latent concepts} $C = \{c_1, c_2, \dots, c_n\}$. One question may relate to one or more latent concepts, and conversely one latent concept may relate to one or more questions. 
Nonetheless, the relationships among these latent concepts are \emph{unknown} (i.e., not predefined). Thus, one important task in solving the knowledge tracing problem is to learn an effective representation (i.e., embedding) of latent concepts and their relationships, as well as to trace the \emph{knowledge state} of a student on latent concepts, i.e., their mastery levels of latent concepts. 

When a student answers questions, the knowledge state of the student is evolving. Each student is thus associated with a sequence of knowledge states $(s_1,\dots,s_{t})$ where $s_1, \dots, s_{t}$ refer to the knowledge states at the time steps $1,\dots,t$, respectively. 
%Further, since latent concepts are not independent and may have relationships among them, they can be naturally represented as a graph. 
%\q{At each time step $t$, a \emph{latent concept graph} $G_t$ can be learned to represent how well a student masters latent concepts individually, as well as their relationships with others.}
In this work, we tackle the KT problem by developing a machine learning model $\mathcal{M}_{\theta}$, paramterized by $\theta$. Concretely, given a question answering sequence $\textbf{x}_t$ and an input question $q_t\in Q$, a KT model $\mathcal{M}_{\theta}$ can trace the knowledge state $s_t$ at each time step $t$ and generate as output the probability $p_t$ of correctly answering the question $q_t$, i.e., $p_t=p(y_t=1|\mathbf{x}_t,s_t,q_t)$.

%% file: Methodology.tex
\section{Deep Graph Memory Network}
\label{sec:methodology}

In this section, we introduce \emph{Deep Graph Memory Network} (DGMN), which has two modules: \emph{attention memory} (AM) and \emph{latent concept graph} (LCG). %The DGMN model fuses information from these two modules to predict the probability of correctly answering a question. 
Figure \ref{fig:Model} shows the model architecture of DGMN.

\subsection{Attention Memory}
\label{sec:am}
The purpose of this module is to learn embedding representations of the latent concepts in a KT task and to track the dynamics of a student's knowledge states. %during training on question answering sequences. (can't be just in training)
%This is achieved through an attention mechanism. 

%The initial step in the AM module is to get a question embedding vector $\mathrm{k_{t}}$. To do so, the one-hot encoding vector of an input question $\mathrm{q_{t}}$ is multiplied with a learnable embedding matrix $\mathrm{\mathbf{A}\in\mathbb{R}^{|Q|\times d_k}}$, where $\mathrm{Q}$ is the unique set of questions. This vector is used to attention the read process from the memory.

\subsubsection{Concept Embedding Memory ($\mathbf{M}^{k}$)}
This memory stores an embedding vector (i.e. a memory slot) for each latent concept. It is represented as a matrix $\mathbf{M}^k\in\mathbb{R}^{N\times d_k}$, where $N$ is the number of latent concepts and $d_k$ is the embedding dimension.

%Let $Q$ be a set of questions in a KT task. 
Given a one-hot encoding vector $\delta(q_t)\in\mathbb{R}^{|Q|}$ of question $q_t$, we multiply it with a learnable embedding matrix $\mathbf{A}\in\mathbb{R}^{|Q|\times d_k}$ to get a question embedding vector $k_{t}\in\mathbb{R}^{d_k}$. Then, the inner product between $k_{t}$ and $\mathbf{M^{k}}$ is supplied to a $\mathrm{Softmax}$ layer to obtain a relevance vector $w_{t}\in \mathbb{R}^N$, as calculated by  %Equation \ref{eq:weightVec} shows the calculation of the relevance vector $\mathrm{w_{t}}$.
\begin{equation}
\label{eq:weightVec}
w_{t}(i)=\mathrm{Softmax}(k_{t}^{\intercal}\mathbf{M}^{k}(i))
\end{equation}

\noindent where $\mathrm{Softmax}(u_{i})=e^{u_{i}}/\sum_{j}e^{u_{j}}$ and $\mathbf{M}^{k}(i)$ is the embedding vector for the $i$-th latent concept. This relevance vector $w_{t}$ reflects the relevance between the current question $q_{t}$ and each latent concept embedding stored in $\mathbf{M}^{k}$. Thus, it is then used to read a student's knowledge state from the concept state memory.

\subsubsection{Concept State Memory ($\mathbf{M}_{t}^{v}$)}
This memory holds the current concept states of a student, represented as a matrix $\mathbf{M}_{t}^{v}\in\mathbb{R}^{N\times d_{v}}$, where $N$ is the number of latent concepts and $d_{v}$ is the state dimension. % for each latent concept.

Given a relevance vector $w_{t}$, we acquire a vector $r_t\in \mathbb{R}^{d_{v}}$ from the $\mathbf{M}_{t}^{v}$ memory using a weighted sum across memory slots as %per Equation \ref{eq:readVec}.
\begin{equation}
\label{eq:readVec}
r_{t}=\sum_{i=1}^{N}w_{t}(i)\mathbf{M}_{t}^{v}(i).
\end{equation}

Here, $r_t$ indicates the concept state information from $\mathbf{M}_{t}^{v}$ relevant to the input question, yet it does not include information about the question itself. To resolve this, we linearly combine it with the question embedding vector $k_{t}\in \mathbb{R}^{d_k}$, and then feed them to a $\mathrm{Tanh}$ layer to calculate a concept summary vector $o_{t}\in\mathbb{R}^{N}$ as %per Equation~\ref{eq:summaryVec}.
\begin{equation}
\label{eq:summaryVec}
o_{t}=\mathrm{Tanh}(\mathbf{W}\left[r_{t},k_{t}\right]+b),
\end{equation}   

\noindent where $\mathrm{Tanh}(u_{i})=(e^{u_{i}}-e^{-u_{i}})/(e^{u_{i}}+e^{-u_{i}})$, $\mathbf{W\in\mathbb{R}^{N\times(d_v+d_k)}}$ is a weight matrix, and $b$ is a bias vector. The concept summary vector $o_{t}$ reflects the information regarding the input question $q_{t}$ and its relevant concept states in $\mathbf{M}_{t}^{v}$.

\subsubsection{Forget Gating Mechanism.} To model a student's forgetting behaviour, we first need to identify questions that are relevant to an input question $q_t$ in terms of its latent concepts. Let $C(q_t)$ be the set of latent concepts underlying a question $q_t$, i.e., for each latent concept $c_i\in C(q_t)$, the relevance weight $w_t(i)$ to $\mathbf{M}^{k}(i)$ is greater than a relevancy threshold $\tau$. A question $q_j$ previously occurring in $\textbf{x}_t$ is \emph{relevant} to $c_i$ iff $c_i\in C(q_j)$. 
Thus, at each time step $t$, we obtain two forgetting features: 
\begin{itemize}
  \item \emph{Time Lapse}: the time lapse $=t-t'$ where $t'$ refers to the most recent time when a question in $\textbf{x}_t$ relevant to the latent concept $c_i$ was observed.
  \item \emph{Trials}: the total number of trials for questions in $\textbf{x}_t$ that are relevant to the latent concept $c_i$.
\end{itemize}

%1) the time delta $\Delta\,t$ since the last time a relevant question to the latent concept was observed, and 2) the total number of trials for questions relevant to the latent concept.

Calculating these forgetting features for $N$ latent concepts at time step $t$ gives us a matrix ${\mathbf{F}_{t}}\in\mathbb{R}^{N\times\,2}$. To get a combined forgetting vector $f_t\in\mathbb{R}^{N}$, we flatten the matrix $\mathbf{F}_{t}$ into an input vector ${f_{in}}\in\mathbb{R}^{2N}$ and send it to a \emph{Tanh} layer as follows:
\begin{equation}
\label{eq:forgetLayer}
{f}_{t}=\mathrm{Tanh}(\mathbf{W}\cdot\,f_{in}+b).
\end{equation}

\noindent where $\mathbf{W}\in\mathbb{R}^{N\times\,2N}$ and $b$ are learnable weight matrix and bias vector respectively.

 To measure the importance of $f_t$ to the input question $q_t$, we perform an element-wise multiplication with the relevance vector $w_{t}\in\mathbb{R}^{N}$ to obtain a forget summary vector ${fs}_t\in\mathbb{R}^{N}$ as
\begin{equation}
\label{eq:forgetVec}
{fs}_{t}=w_{t}\odot\,f_t
\end{equation}

Then, we use a forget gating mechanism to combine information from the concept summary vector $o_t$ and the forget summary vector ${fs}_t$. %This is inspired by the use of gates in LSTMs~\cite{LSTM_97} and gated-RNNs~\cite{GRNN_14} to fuse different sources of information. 
The forget gating mechanism works in two steps. First, a gating vector $gw_t\in\mathbb{R}^N$ is generated by applying a Sigmoid function to a linear combination of vectors $o_t$ and ${fs}_t$. Then, we calculate a combined memory vector $z^{m}_{t}\in\mathbb{R}^N$ based on the gating vector $gw_t$: %The calculations are as follows: 

\begin{equation}
\label{eq:gating}
\begin{aligned}
gw_t=\mathrm{Sigmoid}(\mathbf{W}_{1}\cdot{o_t}+\mathbf{W}_{2}\cdot{fs}_t); \\
z^{m}_t=gw_t\odot\,o_t+(1-gw_t)\odot\,{fs}_t,
\end{aligned}
\end{equation}
%\g{[Explanation] the $z^{m}_t$ is a weighted sum between the memory summary $o_t$ and the forgetting vector $f_t$, thus the weights in the sum must sum to one (that is $gw_t$ and its compliment $1-gw_t$)}

\noindent where $\mathbf{W}_{1}\in\mathbb{R}^{N\times\,N}$ and $\mathbf{W}_{2}\in\mathbb{R}^{N\times\,N}$ are learnable weight parameters, and $\odot$ indicates an element-wise multiplication.
Later, $z^{m}_{t}$ is used along with the information from a latent concept graph (will be introduced in Section~\ref{sec:lcg}) to predicate the probability of correctly answering $q_t$. 

%A block diagram for this gating mechanism is shown in the middle part of Figure~\ref{fig:Model}.

\subsubsection{Memory Update Procedure.}
Updating the state memory $\mathbf{{M}}_{t}^{v}$ is controlled by two signals: \emph{add signal} $a_{t}$ and \emph{erase signal} $e_{t}$.  These signals decide the proportion of information being changed on $\mathbf{{M}}_{t}^{v}$ after observing a new question and its answer from the student. $a_{t}$ decides new information to be written to the memory, while $e_{t}$ decides old information to be removed from the memory.  

Suppose that $(\delta(q_{t}),y_{t})$ is the latest observed question-answer tuple, we embedded it with an embedding matrix $\mathbf{B}\in\mathbb{R}^{2|Q|\times d_{v}}$ to obtain an embedding vector ${v}^q_{t}\in\mathbb{R}^{d_v}$. Then, we concatenate this embedding vector with the summary vector $o_t$ to combine information on the latest question-answer along with the corresponding knowledge state summary. The concatenated update vector $v_t=[{v}^q_t,o_t]\in\mathbb{R}^{(d_v+N)}$ is used to update the $\mathbf{M}_{t}^{v}$ memory using erase and add signals~\cite{NTM14}. The erase signal $e_t\in\mathbb{R}^{d_v}$ is obtained by the following equation:
  \begin{equation}
 \label{eq:forget_gate}
e_{t}=\mathrm{Sigmoid}(\mathbf{W}_{e}\cdot v_{t}+b_{e}).
\end{equation}   

\noindent where $\mathbf{W}_{e}\in\mathbb{R}^{d_v\times\,(d_v+N)}$ is a learnable weight matrix and $b_e$ is a bias vector. Given the erase signal $e_{t}$, the concept state memory is updated from $\mathbf{{M}}_{t}^{v}$ to $\mathbf{\tilde{M}}_{t+1}^{v}$ as
\begin{equation}
\label{eq:Erased_Memory}
\mathbf{\tilde{M}}_{t+1}^{v}(i)=\mathbf{M}_{t}^{v}(i)[1-w_{t}(i)e_{t}].
\end{equation}

The add signal $a_{t}\in\mathbb{R}^{d_v}$ is obtained by the following equation:
\begin{equation}
\label{eq:input_gate}
a_{t}=\mathrm{Tanh}(\mathbf{W}_{a}\cdot v_{t}+b_{a}).
\end{equation}

\noindent where $\mathbf{W}_{a}\in\mathbb{R}^{d_v\times\,(d_v+N)}$ is a learnable weight matrix and $b_a$ is a bias vector. Finally, the concept state memory $\mathbf{M}_{t+1}^{v}$ is calculated as 

\begin{equation}
\label{eq:am_update}
\mathbf{M}_{t+1}^{v}(i)=\mathbf{\tilde{M}}_{t+1}^{v}(i)+w_{t}(i)a_{t}.
\end{equation} 

\subsection{Latent Concept Graph}
\label{sec:lcg}

Although the attention memory module captures temporal dynamics of a student's knowledge state, it cannot represent relationships between latent concepts. For example, observing a knowledge growth for a given latent concept would impact the prediction of answers to related latent concepts. Thus, we learn a graph representation to capture such relationships.

%\subsubsection{LCG Graph Generation.}\label{sec:graph_gen}
%A \emph{latent concept graph} (LCG) is a weighted graph $G=(V,E,\omega)$ which consists of a set of vertices $V$, a set of edges $E \subseteq V \times V$, and a similarity function $\omega$ assigning weights on edges. Each vertex refers to a latent concept, each edge indicate a relationship between two latent concepts, and the similarity function measures the similarity between two latent concepts.

Given an attention memory ($\mathbf{M}^{k}$, $\mathbf{M}_{t}^{v}$), a \emph{latent concept graph} (LCG) over ($\mathbf{M}^{k}$, $\mathbf{M}_{t}^{v}$) is a weighted graph $G=(V,E,\omega)$ where each vertex in $V$ corresponds to a latent concept in $\mathbf{M}^{k}$ and each edge in $E$ is weighted based on the similarity of two latent concepts it connects in $\mathbf{M}_{t}^{v}$. Specifically, we define $\omega:\mathbf{M}_{t}^{v} \times \mathbf{M}_{t}^{v}\rightarrow [0,1]$ such that $\omega(\mathbf{M}_{t}^{v}(i),\mathbf{M}_{t}^{v}(j))\geq\mu$, i.e., the similarity value for the edge between two latent concepts $c_i$ and $c_j$ is no less than $\mu$, where $\mathbf{M}_{t}^{v}(i)$ and $\mathbf{M}_{t}^{v}(j)$ are the concept state vectors for $c_i$ and $c_j$, respectively, and $\mu\in [0,1]$ is a similarity threshold.

%\subsubsection{Distilling Information from LCG Graph.}
To extract useful information from a latent concept graph, we employ graph convolutional networks (GCNs)~\cite{GCN_NIPS2015,kipf_GCN_17}. This is achieved by applying convolution over the neighbours of each vertex in the graph. The input of a GCN layer includes 1) the latent concept embedding memory $\mathbf{M^{k}}\in\mathbb{R}^{N\times d_{k}}$ in our case, and 2) the graph adjacency matrix with self-loops $\hat{\mathbf{A}}=\mathbf{A}+\mathbf{I}$ and $\hat{\mathbf{A}}\in\mathbb{R}^{N\,\times\,N}$, where $\mathbf{I}$ is the identity matrix, and 3) the diagonal degree matrix $\hat{\mathbf{D}}\in\mathbb{R}^{N\,\times\,N}$ containing the degree information for each vertex in $\hat{\mathbf{A}}$. The propagation rule of a GCN layer is calculated through a convolution process \cite{kipf_GCN_17}:

\begin{equation}
\label{eq:gcn}
\mathbf{H}^{i+1}=ReLU({\hat{\mathbf{D}}^{-\frac{1}{2}}}{\hat{\mathbf{A}}}{\hat{\mathbf{D}}^{-\frac{1}{2}}}{\mathbf{H}^{i}}{\mathbf{W}^{i}})
\end{equation}

%\begin{equation}
%\label{eq:gcn}
%\mathbf{H}^{i+1}=ReLU((\hat{\mathbf{D}}^{-1/2}\dot\,\hat{\mathbf{A}}\,\dot\,\hat{\mathbf{D}}^{-1/2}\dot\,\mathbf{H}^{i})\dot\,\mathbf{W}^{i+1} + b^{i+1})
%\end{equation}

\noindent where $\mathrm{ReLU}(u)=max(0,u)$ is a non-linear activation function, $\mathbf{W}^{i}$ is a learnable weight matrix. ${\mathbf{H}^i}\in\mathbb{R}^{N\,\times\,d_k}$ is an input feature matrix which is the output of the previous GCN layer, and $\mathbf{H}^0=\mathbf{M}^{k}$ for the first GCN layer. 

Finally, we calculate the \emph{graph summary vector} $z^{g}_{t}\in\mathbb{R}^{N}$, which attends the GCN output to the current question $q_t$ using the relevance vector $w_t$ 
\begin{equation}
\label{eq:gsummaryVec}
z^{g}_{t}=\mathrm{Tanh}(\mathbf{W}(\sum_{i=1}^{N}w_{t}(i)\mathbf{H}(i))^{\intercal}+b),
\end{equation} 

\noindent where $\mathbf{W}\in\mathbb{R}^{N\times{d_k}}$ is a learnable weight matrix, $\mathbf{H}$ is the output matrix of the final $\mathrm{GCN}$ layer in our model, and $b$ is a bias vector.

\subsection{Answer Prediction}
\label{sec:answer_pred}

To predict the probability of correctly answering an input question $q_t$, we concatenate the combined memory vector $z^{m}_{t}$ from the AM module with the graph summary vector $z^{g}_{t}$ from the LCG module and feed them to a fully connected layer with Sigmoid activation. Then, we multiply the output logits vector with the one-hot encoding vector $\delta(q_t)$ of question $q_t$ as
\begin{equation}
\label{eq:Final_Prob}
p_{t}=\mathrm{Sigmoid}(\mathbf{W}\left[z^{m}_{t},z^{g}_{t}\right]+b)\cdot\delta(q_t).
 \end{equation} 

\noindent where $\mathbf{W}\in\mathbb{R}^{|Q|\times{2N}}$ is a learnable weight matrix, $b$ is a bias vector, and $p_t$ is a scalar predicting the correct answer probability for $q_t$.

\subsection{Model Optimization}
\label{sec:model_opt}

%\subsubsection{Question Ranking.} 
Different questions may have different impacts on a student's knowledge state. We observe that the impact of a question largely depends on its latent concepts and the importance of these concepts in mastering a learning subject. Consider the dataset \emph{Statics2011} for example, which was collected from an undergraduate mechanical engineering course (will be further discussed in Section \ref{sec:datasets}). As ``Forces'' is a fundamental concept in the course, correctly answering questions about this concept can significantly help improve the performance on the other concepts.

 %are  $\gamma(q_t)\in[0,1]$ 

Following this observation, we exploit a question ranking technique during model training. Given a training mini-batch $\hat{D}^{train}=\{(q_j,y_j)\}^{m}_{j=1}$, the ranks for questions in $\hat{D}^{train}$ are calculated as follows:
\begin{equation}
    \label{eq:exercise_rank}
    \begin{aligned}
    \gamma(q_j) = \sum_{i=1}^{N}w_{j}(i)\,\mathbf{D}(i,i)\\
%    \g{\gamma_{q_j} = \mathrm{minmax\_norm}(\overline{\gamma}_{q_j},\Gamma_{\hat{D}^{train}})}
    \end{aligned}
 \end{equation}
\noindent where $w_j$ is the relevance vector of $q_j$ from the attention memory module, $\mathbf{D}(i,i)$ refers to the degree value of the $i$-th latent concept $c_i$ from the latent concept graph, and $N$ refers to the total number of latent concepts in the KT task.

Let $\Gamma=\{\gamma(q_j)|q_j\in \hat{D}^{train}\}$ be the set of question ranks for $\hat{D}^{train}$. Then, we apply min-max normalization to ensure that the ranks in $\Gamma$ are normalised s.t. each question $q_j$ is associated with a normalised rank $\overline{\gamma}(q_j)\in [0,1]$. Then, given a training mini-batch $\hat{D}^{train}$, our model aims to minimize the following objective function:
\begin{equation}
\label{eq:loss_func}
{L}=\frac{1}{|\hat{D}^{train}|}\sum_{i=1}^{|\hat{D}^{train}|}\,\overline{\gamma}(q_{i})\,\cdot\,l_i.
\end{equation}

\noindent where $l_i=-(y_i\log(p_i)+(1-y_i)\log(1-p_i))$ is the cross-entropy loss for predicting the answer $y_{i}$ of question $q_{i}$. 

%\subsubsection{Training Algorithm.}
The training process of our proposed DGMN model is described in Algorithm \ref{alg:DGMN}.

\begin{algorithm}
\caption{Training DGMN}\label{alg:DGMN}
\begin{algorithmic}[1]
\Initialize{
    Weights, bias parameters and embedding matrices}\newline
\textbf{Begin:}
\For{epoch$\,\in\,1,\dots,n$}
  \For{mini-batch$\,\in\,1,\dots,m$}
  \ForAll{sample$\,\in\,$mini-batch}
    \State{Embed an input question $q_{t}$} %^{\intercal}\times\mathbf{A}$}
    \State{Get relevance vector ${w_t}$ as per Eq.~\ref{eq:weightVec} }
    \State{Get read vector ${r_t}$ as per Eq.~\ref{eq:readVec}}
    \State{Get concept summary vector ${o_t}$ as per Eq.~\ref{eq:summaryVec}}
    \State{Get forgetting summary vector ${fs}_{t}$ as per Eg.~\ref{eq:forgetVec}}
    \State{Get combined memory vector ${z^{m}_{t}}$ as per Eq.~\ref{eq:gating}}
    \State{Get graph summary vector $\mathrm{{z^{g}}_{t}}$ as per Eq.~\ref{eq:gsummaryVec}}
    \State{Compute the probability $\mathrm{p_t}$ as per Eq.\ref{eq:Final_Prob}}
    \State{Perform update on attention memory using
    Eq.~\ref{eq:am_update}}
    \EndFor
    \State{Get question ranking in the mini-batch using Eq.~\ref{eq:exercise_rank}}
    \State{Perform Adam~\cite{santoro16} w.r.t. Eq.~\ref{eq:loss_func}}
    \State{Update the LCG graph}
  \EndFor
\EndFor
\end{algorithmic}
\end{algorithm}

%% file: Experiments.tex
\section{Experiments}
\label{sec:exp}
In this section, we present the experimental design for evaluating our proposed DGMN model, aiming to answer the following research questions:

\begin{itemize}
\item[\emph{RQ1}:] How does our proposed DGMN model perform against the state-of-the-art KT models?
\item[\emph{RQ2}:] How different components (i.e., forget gating mechanism, latent concept graph and question ranking) in the proposed DGMN model affect on its performance?
\item[\emph{RQ3}:] How effectively can a learned latent concept graph reflect latent concepts and their relationships?
\item[\emph{RQ4}:] How does modelling forgetting over multiple concepts in our model compare to other deep learning models assuming only a single concept?
\end{itemize}

\subsection{Datasets}
\label{sec:datasets}
In this section, we describe the datasets used in our experiments. Table \ref{tbl:datasets} summarizes some statistics for each dataset.   
\begin{itemize}
\item \textbf{ASSISTments2009\footnote{ASSISTments2009:\url{  https://sites.google.com/site/assistmentsdata/home/assistment-2009-2010-data/skill-builder-data-2009-2010}}:} This dataset includes questions from school math topics. It was generated using the ASSISTments online education website \footnote{https://www.assistments.org/} between years $2009-2010$. It has $110$ unique questions and $4,151$ of participating students with a total of $325,637$ exercises (i.e., question and answer pairs). There are $110$ of latent concepts (i.e., mathematical concepts) in this dataset.

\item \textbf{Statics2011\footnote{Statics2011:\url{https://pslcdatashop.web.cmu.edu/
DatasetInfo?datasetId=507}}:} This dataset contains questions gathered from an engineering course conducted by the Carnegie Mellon University during $2011$ academic year. It has a total number of $333$ participating students answering a unique set of $1,223$ questions. The total exercise set has $189,297$ exercises. There are $85$ of latent concepts in this dataset.

\item \textbf{Synthetic-5\footnote{Synthetic-5:\url{https://github.com/chrispiech/DeepKnowledgeTracing/tree/master/data/synthetic}}:} This dataset was generated by the authors of DKT~\cite{DKT2015}, where they simulated a learning experience for $4,000$ student agents answering $50$ unique questions. This has led to generating a total of $200K$ exercises. There are $5$ latent concepts in this dataset.

\item \textbf{Kddcup2010\footnote{Kddcup2010:\url{https://pslcdatashop.web.cmu.edu/KDDCup/downloads.jsp}}:} This dataset is based on an algebra course conducted on the cognitive algebra tutor system~\cite{kdd2010} between $2005-2006$. It consists of $436$ unique questions being answered by $575$ students, which gives an exercise set of size $607,026$. There are $112$ latent concepts in this dataset.
\end{itemize}

\begin{table}
  \caption{Statistics of datasets}
  \label{tbl:datasets}\vspace*{-0.3cm}
  \resizebox{\columnwidth}{!}{
  \begin{tabular}{lcccc}
    \toprule
    \hspace*{0cm}Dataset&\#Students&\#Questions&\#Exercises&\#Concepts \\
    \midrule
    ASSISTments2009&$4,151$&$110$&$325,637$&$110$\\
   Statics2011&$335$&$1,362$&$190,923$&$85$\\
    Synthetic-5&$4,000$&$50$&$200,000$&$5$\\
    Kddcup2010&$575$&$436$&$607,026$&$112$\\
  \bottomrule
\end{tabular}
}
\end{table}

\subsection{Experimental Setup}
 To answer the previously mentioned research questions, our experimental design involves four experiments as follows.

 \begin{table*}
  \caption{The average AUC results for comparing DGMN with the state-of-the-art KT models over all the datasets.}\vspace{-0.3cm}
  \label{tbl:AUC}
   \begin{adjustbox}{max width=\textwidth}
  \begin{tabular}{l|cccc|cccc}
    \toprule
     \multirow{2}{*}{Model}& \multirow{2}{*}{Forget}&\multirow{2}{*}{Graph} &\multirow{2}{*}{Rank}&\multirow{2}{*}{Knowledge State}&\multicolumn{4}{c}{Dataset}\\\cline{6-9}
   &&&&&ASSISTments2009&Statics2011&Synthetic-5&Kddcup2010\\
      \midrule
      GKT~\cite{GKT19}&$\times$&\checkmark&$\times$&GNN&$72.3\pm0.02$&$73.4\pm0.03$&$74.2\pm0.01$&$76.9\pm0.01$\\
      DKT+forget~\cite{DKT_Forget}&over one latent concept&$\times$&$\times$&RNN&$73.2\pm0.02$&$74.5\pm0.01$&$75.1\pm0.03$&$79.0\pm0.01$\\
    DKVMN~\cite{DKVMN17}&$\times$&$\times$&$\times$&Key-value memory&$81.6\pm0.03$&$82.8\pm0.02$&$82.7\pm0.01$&$79.8\pm0.02$\\
    SAKT~\cite{SAKT}&$\times$&$\times$&$\times$&Attention mechanism&$83.7\pm0.02$&$84.2\pm0.01$&$81.9\pm0.03$&$80.4\pm0.02$\\
   AKT~\cite{AKT20}&$\checkmark$&$\times$&$\times$&Attention mechanism&$84.1\pm0.03$&$85.0\pm0.02$&$83.6\pm0.03$&$81.5\pm0.02$\\\hline
    DGMN (ours)&over multiple latent concepts&\checkmark&\checkmark&Key-value memory&$\mathbf{86.1\pm0.01}$&$\mathbf{86.4\pm0.02}$&$\mathbf{85.9\pm0.03}$&$\mathbf{83.4\pm0.01}$\\
    \bottomrule
\end{tabular}
\end{adjustbox}
\end{table*}
 \begin{figure*}[t!]
\includegraphics[width=1\textwidth]{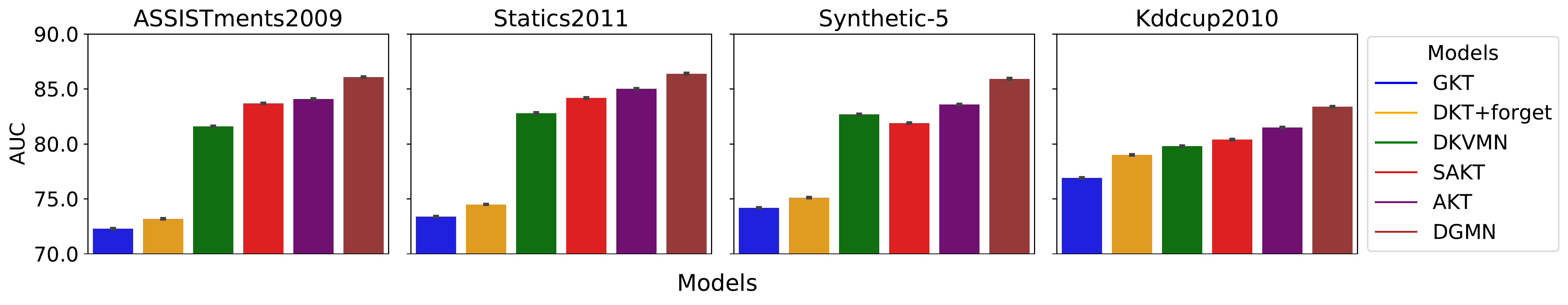}\vspace{-0.3cm}
\caption{The average AUC values comparing DGMN with the state-of-the-art KT models over all the datasets (best in colors).}
\label{fig:AUC_res}\vspace{-0.1cm}
\end{figure*}

\subsubsection{Comparison with State-of-the-art KT Models.} This experiment answers the research question \emph{RQ1}. This is done by comparing the performance of our DGMN model with the state-of-the-art KT models on four well-established datasets in the KT literature. %Section \ref{sec:datasets} describes the characteristics of these datasets in detail.  
The KT models for comparison include:
\begin{itemize}
\item \textbf{DKT+forget}~\cite{DKT_Forget}:
This model extends the DKT model~\cite{DKT2015} by adding forgetting features calculated from a question answering sequence. 
\item \textbf{DKVMN}~\cite{DKVMN17}:
This model uses a key-value memory to track the knowledge state of a student across latent concepts. 
\item \textbf{GKT}~\cite{GKT19}:
This model uses a graph neural network to extract information that captures dependencies across questions. 
\item \textbf{SAKT}~\cite{SAKT}: 
This model follows an attention mechanism~\cite{BahdanauCB14} to weight the importance of previously answered questions in a sequence for predicting the latest one.
\item \textbf{AKT}~\cite{AKT20}: This model combines an attention model with Rasch model-based embeddings, which exponentially decays attention weights w.r.t. the distance of questions in a sequence in order to account for student's forgetting effect.
\end{itemize}

We use the best parameter configurations as suggested in the original work for these models. For all the models including our DGMN model, we execute five independent runs and report the averaged results. We also perform the statistical significance \emph{t-test} on the results with \emph{p-value}$\mathrm{=0.05}$. %We visualize the ROC curve for each comparison model and for each dataset. To summarize the results, we report the average AUC metric for each model. 

\subsubsection{Ablation Study of DGMN}
To answer the research question \emph{RQ2}, we conduct an ablation study. We compare different variants of our DGMN model to evaluate the impact of the LCG graph module, forget gating mechanism, and question ranking technique on the performance of DGMN. These variants are as follows:
\begin{itemize}
    \item \textbf{DGMN-Basic}: This variant only includes a basic KV memory without using the LCG graph module, forget gating mechanism, and question ranking technique.
    \item \textbf{DGMN-NoForget}: This variant is obtained by removing the forget gating mechanism and question ranking technique from DGMN. The purpose of this variant is to evaluate the impact of using latent concept dependencies extracted from the LCG graph on the performance of DGMN.
    \item \textbf{DGMN-NoGraph}: This variant is obtained by removing the LCG graph module and question ranking technique from DGMN. Note that, the question ranking technique has to be removed in this variant because it depends on the degree information from the LCG graph. The purpose of this variant is to evaluate the impact of the forget gating mechanism on the performance of DGMN.
    \item \textbf{DGMN-NoRank}: This variant includes all the components of DGMN except for the question ranking technique mentioned in Section~\ref{sec:model_opt}. This variant is to evaluate the impact of question ranking on the performance of DGMN.
\end{itemize}

%To assess the impact of the LCG graph, we compare results for the DGMN-G variant with the DGMN-B variant. For the impact of the forget gate, we compare the results of the DGMN-F variant with the DGMN-B variant. For the impact of the question ranking technique, we compare the DGMN-R variant with the DGMN model itself.

\subsubsection{Analysis of LCG Graphs.} To answer the research question \emph{RQ3}, we visualize the LCG graphs learned from two datasets -- \emph{ASSISTments2009} and \emph{Statics2011}. We label the nodes with their corresponding latent concept numbers, and the edges between nodes reflect relationships. In addition to this, we cluster the nodes that are close to each other in a node embedding space. Different clusters are highlighted using different colors. We analyse how these clusters reflect semantic relatedness between latent concepts. %\q{These clusters are compared against the ground truth provided in the datasets. (?how)} 

\subsubsection{Analysing of Forget Modelling.} To answer the research question \emph{RQ4}, this experiment aims to discuss the impact of modelling a student's forgetting behaviours over multiple latent concepts in comparison to doing it assuming only a single latent concept. To achieve this, we compare the probabilities of correctly answering questions between DGMN and DKT+forget models using a sample question sequence from the \emph{ASSISTments2009} dataset. We visualize their differences using a heatmap (will be depicted in Figure \ref{fig:forget_heat}).

\subsection{Implementation Details}
\label{sec:implementation}
For each dataset, we divide it into training and testing sets using a split ratio of $\mathrm{(70\%,30\%)}$ for training and testing, respectively. We use a 5-fold cross validation strategy in our experiments. The attention memory matrices ($\mathbf{M}^{k}$ and $\mathbf{M}_{t}^{v}$) and embedding matrices ($\mathbf{A}$ and $\mathbf{B}$) are initialized using a zero-mean random Gaussian distribution $N(0,\sigma)$. For parameters of the neural layers, we use \emph{Glorot} uniform random initialization \cite{glorot2010} for a faster convergence. %\q{The number of latent concepts $N$ is determined as per ground truth in each dataset}. 
An empirical evaluation on validation data (i.e., $20\%$ of training data) is conducted to determine the hyper-parameters of our model, including: the memory slot dimensions $d_k=50$ and $d_v=100$, and the relevancy threshold $\tau=0.8$ for forgetting features calculation. The similarity function $\omega$ is chosen to be \emph{cosine similarity} \cite{sim20} and the similarity threshold $\mu=0.25$. We optimize the model’s parameters using \emph{Adam} gradient decent algorithm \cite{santoro16}.

%% file: Results.tex
\section{RESULTS AND DISCUSSION}
\label{sec:res}

In this section, we present results of our experiments and discuss the findings from these experiments.

\subsection{Prediction Performance}
This experiment compares the performance of the DGMN model with the state-of-the-art KT models on four KT datasets.  Table~\ref{tbl:AUC} reports the average AUC results and Figure~\ref{fig:AUC_res} depicts average AUC values using bar plots.

\begin{comment}
\begin{table}[hbt!]
  \caption{The average AUC with standard deviation results for KT models over all datasets.} 
  \resizebox{\columnwidth}{!}{%
  \begin{tabular}{lcccl}
    \toprule
    Dataset&ASSISTments2009&Statics2011&Synthetic-5&Kddcup2010\\
    \midrule
    GKT&$72.3\pm0.02$&$73.4\pm0.03$&$74.2\pm0.01$&$76.9\pm0.01$\\
    DKT+forget&$73.2\pm0.02$&$74.5\pm0.01$&$75.1\pm0.03$&$79.0\pm0.01$\\
    DKVMN&$81.6\pm0.03$&$82.8\pm0.02$&$82.7\pm0.01$&$79.8\pm0.02$\\
    SAKT&$84.8\pm0.03$&$85.3\pm0.01$&$83.2\pm0.02$&$81.7\pm0.01$\\\hline
    DGMN (Ours)&$\mathbf{86.1\pm0.01}$&$\mathbf{86.4\pm0.02}$&$\mathbf{85.9\pm0.03}$&$\mathbf{83.4\pm0.01}$\\
  \hline
\end{tabular}
}
\label{tbl:AUC}
\end{table}
\end{comment}
\begin{comment}

\begin{table}
  \caption{Ablation study comparing four variants of the DGMN model.}
  \label{tbl:ablation}
   \resizebox{\columnwidth}{!}{
  \begin{tabular}{lcccc}
    \toprule
    Dataset&DGMN-F&DGMN-G&DGMN-R&DGMN-B\\
    \midrule
    ASSISTments2009&$82.8\pm0.01$& $85.2\pm0.02$&$85.7\pm0.02$&$81.9\pm0.01$\\
    Statics2011&$83.9\pm0.03$& $85.5\pm0.01$&$85.9\pm0.01$&$82.9\pm0.03$\\
    Synthetic-5& $83.8\pm0.04$& $84.7\pm0.02$&$85.1\pm0.02$&$82.9\pm0.01$\\
    Kddcup2010& $80.7\pm0.03$& $82.4\pm0.01$&$82.7\pm0.03$&$80.0\pm0.02$\\
  \bottomrule
\end{tabular}
}
\end{table}
\end{comment}

We observe that DGMN outperforms all the other models on all the datasets and achieves an AUC improvement over the nearest model (i.e. {AKT}) by a margin of  $\mathrm{2.0\%}$, $\mathrm{1.4\%}$, $\mathrm{2.3\%}$, and $\mathrm{1.9\%}$ on the datasets \emph{ASSISTments2009}, \emph{Statics2011}, \emph{Synthetic-5}, and \emph{Kddcup2010}, respectively. These results are statistically significant, i.e., \emph{p-value}$\mathrm{<0.05}$, based on the \emph{t-test} conducted in different runs. Notice that, DGMN and AKT both obtain the lowest AUC values on \emph{Kddcup2010} among all the databases. This is due to the complexity that this dataset adds to the KT benchmarks as it has the largest number of exercises and latent concepts among all the datasets.

The results highlight the performance advantage of DGMN by utilizing graph features from LCGs and forgetting features from attention memory, in comparison to only forgetting features as in DKT-forget, only a key-value memory without any forgetting features as in DKVMN, and only graph features as in GKT.

\vspace*{-0.1cm}

\subsection{Ablation Study}
Table~\ref{tbl:ablation} summarizes the average AUC results for different variants of the DGMN model. From Table~\ref{tbl:ablation}, we have the following findings: \vspace*{-0.2cm}
\begin{itemize}
    \item[(1)] The impact of the LCG graph module can be observed by comparing DGMN-Basic and DGMN-NoForget. A statistically significant (p-value < 0.05) performance enhancement by a margin of $\mathrm{0.9\%}$, $\mathrm{1.0\%}$, $\mathrm{0.9\%}$, and $\mathrm{0.7\%}$ is achieved for \emph{ASSISTments2009}, \emph{Statics2011}, \emph{Synthetic-5} and \emph{Kddcup2010}, respectively.
    \item[(2)] The impact of the forget gating mechanism can be observed through comparing DGMN-Basic and DGMN-NoGraph. DGMN-NoGraph achieves a statistically significant performance enhancement upon DGMN-Basic by a margin of $\mathrm{3.3\%}$, $\mathrm{2.6\%}$, $\mathrm{1.8\%}$, and $\mathrm{2.4\%}$ for \emph{ASSISTments2009}, \emph{Statics2011}, \emph{Synthetic-5} and \emph{Kddcup2010}, respectively. 
    \item[(3)] The impact of the question ranking technique can be observed by comparing DGMN-NoRank with DGMN. A significant performance margin of $\mathrm{0.4\%}$, $\mathrm{0.5\%}$, $\mathrm{0.8\%}$, and $\mathrm{0.7\%}$ exists between these two models for \emph{ASSISTments2009}, \emph{Statics2011}, \emph{Synthetic-5}, and \emph{Kddcup2010}, respectively.
\end{itemize} 

\begin{table*}
  \caption{The Average AUC results for comparing different variants of DGMN over all the datasets.}\vspace*{-0.3cm}
  \label{tbl:ablation}
   \begin{adjustbox}{max width=\textwidth}
  \begin{tabular}{l|cccc|cccc}
    \toprule
     \multirow{2}{*}{Model\hspace*{2cm}}& \multicolumn{4}{c|}{Component}& \multicolumn{4}{c}{Dataset}\\\cline{2-9}
     &Forget&Graph&Rank&KV Memory&ASSISTments2009&Statics2011&Synthetic-5&Kddcup2010\\
      \midrule
   DGMN&\checkmark&\checkmark&\checkmark&\checkmark&$\mathbf{86.1\pm0.01}$&$\mathbf{86.4\pm0.02}$&$\mathbf{85.9\pm0.03}$&$\mathbf{83.4\pm0.01}$\\\hline
   DGMN-NoForget &$\times$&\checkmark&$\times$&\checkmark&$82.8\pm0.01$&$83.9\pm0.03$& $83.8\pm0.04$& $80.7\pm0.03$\\
   DGMN-NoGraph&\checkmark&$\times$&$\times$&\checkmark&$85.2\pm0.02$&$85.5\pm0.01$&$84.7\pm0.02$&$82.4\pm0.01$\\
   DGMN-NoRank&\checkmark&\checkmark&$\times$&\checkmark&$85.7\pm0.02$&$85.9\pm0.01$&$85.1\pm0.02$&$82.7\pm0.03$\\
   DGMN-Basic&$\times$&$\times$&$\times$&\checkmark&$81.9\pm0.01$&$82.9\pm0.03$&$82.9\pm0.01$&$80.0\pm0.02$\\
   \bottomrule
\end{tabular}
\end{adjustbox}
\end{table*}
\begin{figure*}[t!]
\includegraphics[width=1\textwidth, height=3cm]{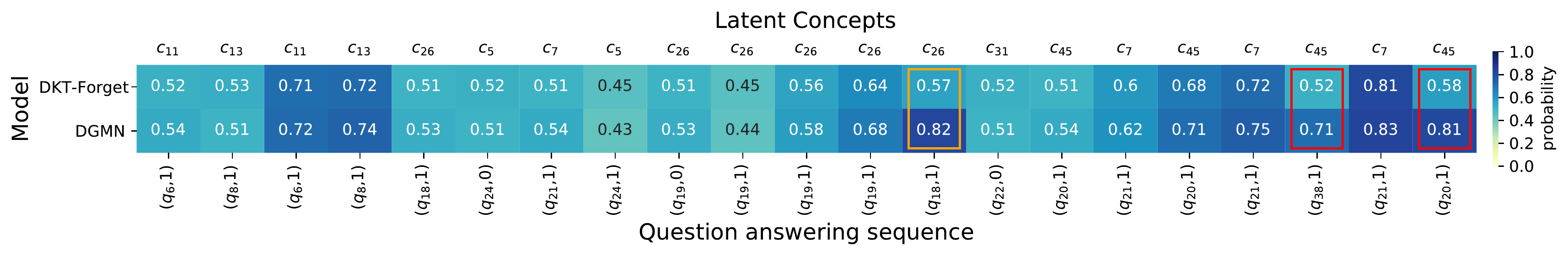}\vspace{-0.3cm}
\caption{A heatmap comparing the probabilities of correctly answering questions, predicted by the DKT+forget and DGMN models, where the question answering sequence is taken from the \emph{ASSISTments2009} dataset (best in colors).}
\label{fig:forget_heat}
\end{figure*}

\subsection{Analysis of LCG Graphs}\label{subsec:graph}
In this experiment, we visualize the LCG graphs learned by DGMN from two of the datasets: \emph{ASSISTments2009} and \emph{Statics2011} in Figure~\ref{fig:learned_graphs}. We apply the following principles to construct a LCG graph: 1) assigning a unique number for each latent concept; % based on the information provided in each dataset; 
2) clustering similar latent concepts based on their embeddings in the embedding space of $\mathbf{M}^k$; 3) highlighting latent concepts in the same cluster with the same color and edges with the colors of their incident nodes with higher degrees. %; and 4) setting the sizes of nodes proportionally to their degrees.
% demonstrates two LCG graphs learned by DGMN over the \emph{ASSISTments2009} and \emph{Statics2011} datasets. 

We notice that, latent concepts that are close in the embedding space have closely related meanings in these datasets. This reflects the effectiveness of embedding representations for latent concepts in DGMN. In the LCG graph for \emph{ASSISTments2009} in Figure~\ref{fig:learned_graphs}.a, there are $10$ clusters of latent concepts. In the blue cluster that represents latent concepts for solving system of equations, one can see that concepts with the numbers $\{26,65,97\}$ are close to each other and their text descriptions are ``\emph{Equation solving two or fewer steps}'', ``\emph{Scientific notation}'', and ``\emph{Choose an equation from given information}'', respectively. Similarly, in the LCG graph for \emph{Statics2011} in Figure~\ref{fig:learned_graphs}.b, there are $7$ clusters of latent concepts. In the purple cluster that represents latent concepts about forces and their mechanical laws, the concepts with the numbers $\{17,82,35\}$ are close to each other, and they have the descriptions ``\emph{recognize concurrent forces}'', ``\emph{rotation of forces}'', and ``\emph{moving force perpendicular to line of action}'', respectively.

These LCG graphs also reflect the relationships between latent concepts. In the LCG graph for \emph{ASSISTments2009} in Figure~\ref{fig:learned_graphs}.a, the edges $\{(93,77),(87,61),$ $(39,92)\}$ represent relationships between the latent concepts (``\emph{rotation}'', ``\emph{3D figures}''), (``\emph{algebraic solving}'', ``\emph{greatest common factor}''), and (``\emph{box and whisker}'', ``\emph{range}''), respectively. It can be observed that the blue cluster that represents ``\emph{solving system of equations}'' is a fundamental one for many other clusters such as ``\emph{geometric theorems}'', ``\emph{statistics}'', or ``\emph{probability}''. Similarly, in the LCG graph for \emph{Statics2011} in Figure~\ref{fig:learned_graphs}.b, the edges $\{(6,42),(12,11),(27,46)\}$ reflect relationships between the latent concepts (``\emph{distinguish rotation translation}'', ``\emph{rotation sense of force}''), (``\emph{represent interaction on contacting body}'', ``\emph{draw force on body}''), and (``\emph{static problem force and moment}'', ``\emph{moment sign sense relation}''), respectively. The purple cluster is densely connected to all the other clusters as it contains fundamental latent concepts such as ``\emph{forces}'', while the other clusters represent operations related to it such as ``\emph{rotations}'', ``\emph{moments}'', or ``\emph{motion}''.

%Finally, from Figure~\ref{fig:learned_graphs}, we can see that these LCG graphs reflect the fact that latent concepts in the same cluster are often closely related to each other. 

\subsection{Analysis of Forget Modelling}
To assess the effectiveness of our forgetting modelling, we examine the probabilities of predicting correct answers by the DGMN and DKT+forget models. Figure~\ref{fig:forget_heat} shows a heatmap in which the question answering sequence is taken from the dataset \emph{ASSISTments2009}. We spot two cases that highlight the differences in forgetting modelling over multiple latent concepts considering their mutual relationships and over a single latent concept. For clarity, these two cases are marked with the orange and red rectangles in Figure~\ref{fig:forget_heat}, respectively. 
\begin{itemize}
\item For the first case, DKT+forget produces a lower prediction probability for $q_{18}$ because forgetting features assuming one latent concept are \emph{Time Lapse}=8 and 
\emph{Trials}=2. % that leads to predicting the occurrence of forgetting. 
In contrast, DGMN produces a higher prediction probability for $q_{18}$ as DGMN is able to detect that $q_{19}$ and $q_{18}$ relate to the same latent concept $c_{26}$, thereby yielding a lower forgetting chance with \emph{Time Lapse}=1 and \emph{Trials}=6. 
\item For the second case, DKT+forget predicts lower probabilities for questions $q_{38}$ and $q_{20}$ because for the former question it was the first time to appear in the sequence and the latter question indicates a higher forgetting chance due to \emph{Time Lapse}=4 and \emph{Trials=3}. In contract, DGMN predicts higher probabilities because given that the questions $q_{38}$ and $q_{20}$ relate to the same concept $c_{45}$, this yields a lower forgetting chance with \emph{Time Lapse}=2 and \emph{Trials}=4. 
\end{itemize}These findings support the effectiveness of modelling forgetting features over multiple concepts while considering their relationships captured by the LCG module.

%% file: RelatedWork.tex
\section{Related Work}
\label{sec:rl}
%In this section, we discuss the research areas related to the present work.
Generally, there are three main directions of research relating to knowledge tracing: memory-augmented neural networks for capturing knowledge states, graph neural networks for representing relationship features, and modelling of forgetting behaviours. Table \ref{tbl:AUC} summaries the state-of-the-art KT methods in these areas. Below, we review them accordingly.

% including 1) memory-augmented neural networks, 2) graph neural networks, 3) knowledge tracing methods, and 4) modelling student's forgetting behaviour. \g{We contrast the key design aspects including the way of modelling temporal dynamics, using graph representations, and modelling a student's forgetting behavior across the state-of-the-art KT models in Table~\ref{tbl:aspect}.}

\begin{figure}[t!]
\centering

\subfloat[]{
	\label{subfig:correct}
	\includegraphics[height=4cm, width=0.9\columnwidth]{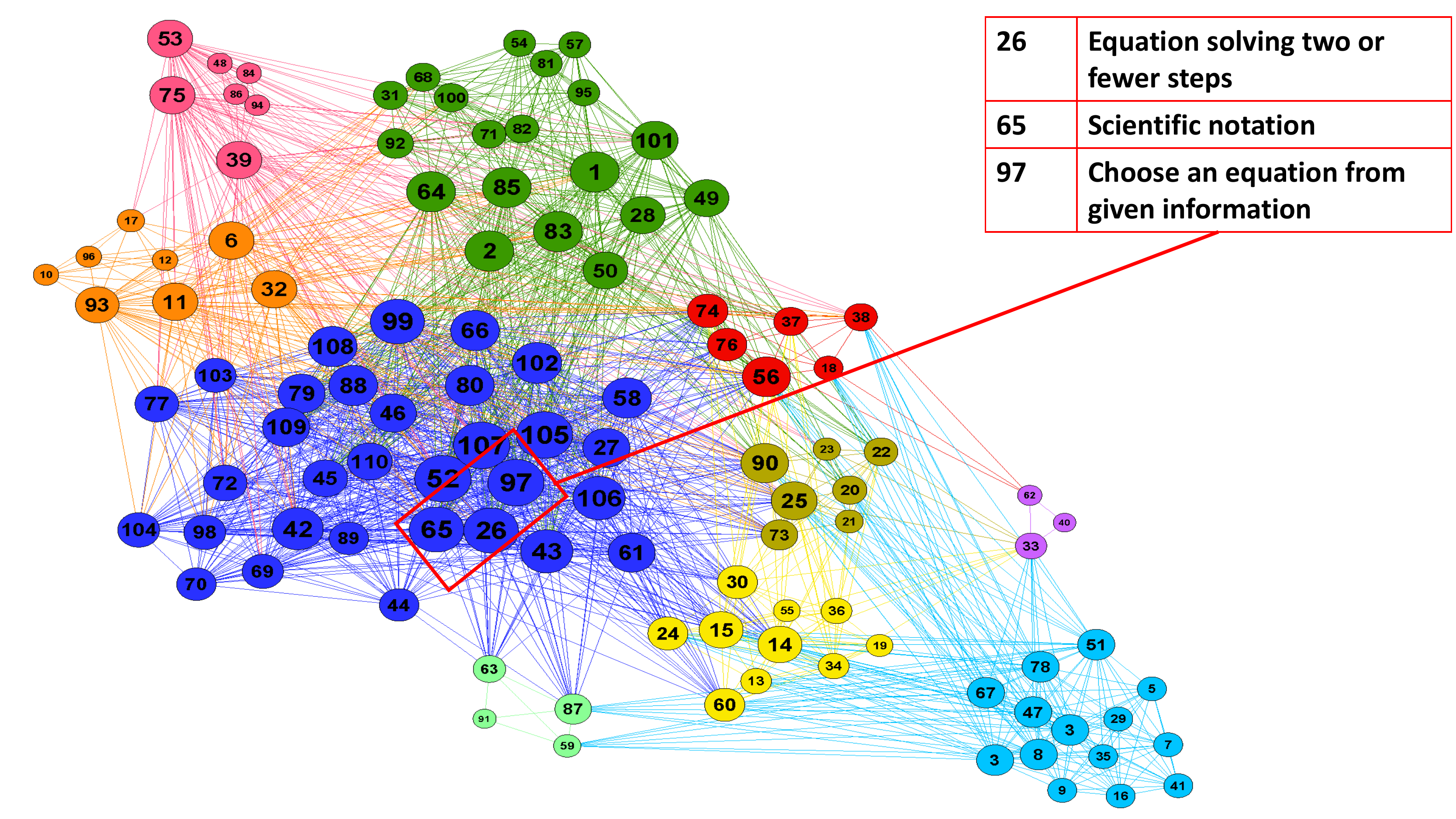} }
	
\subfloat[]{
	\label{subfig:notwhitelight}
	\includegraphics[height=4cm, width=0.9\columnwidth]{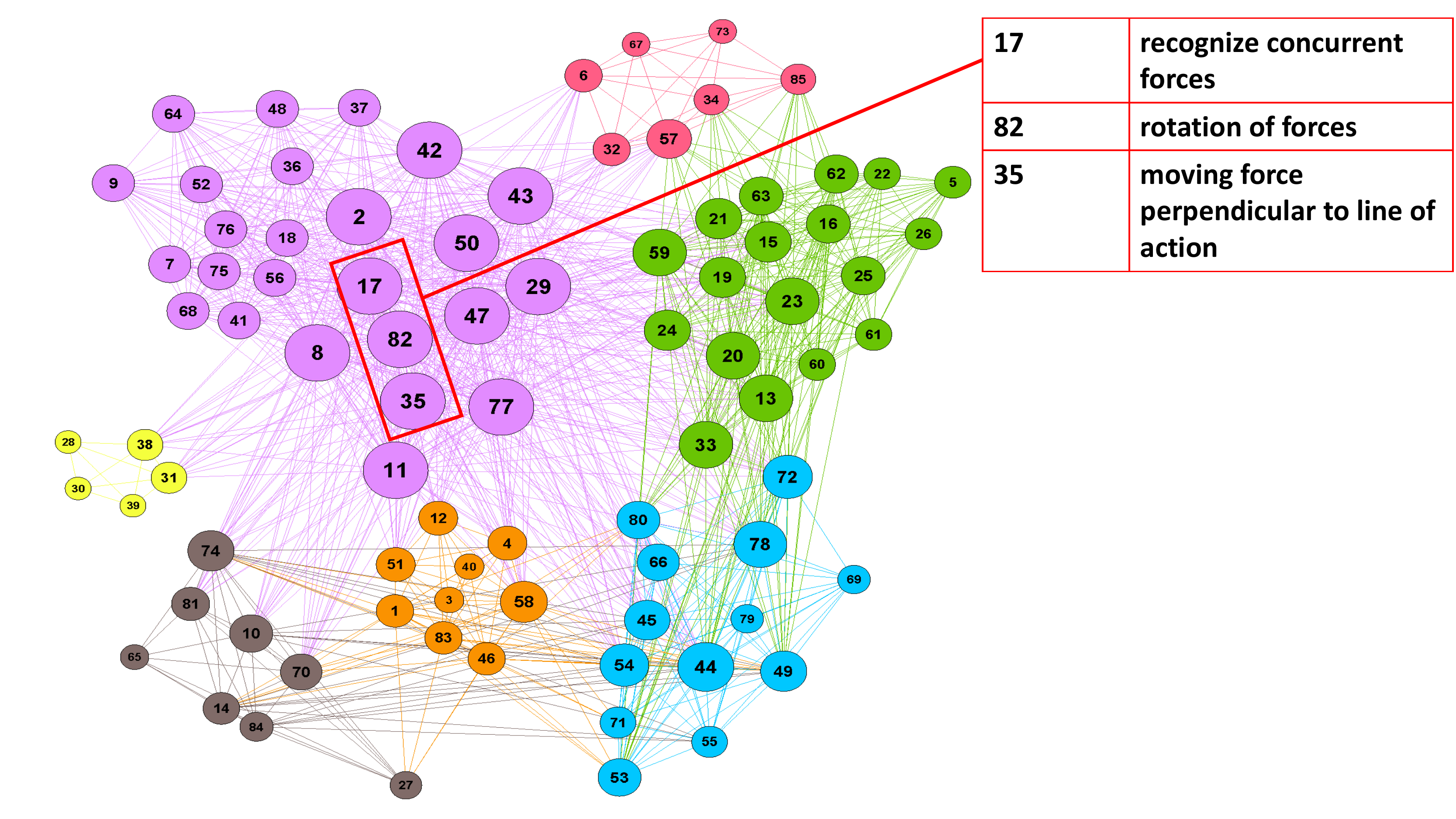} }\vspace{-0.3cm}
	
\caption{Learned LCG graphs in two datasets: (a) \emph{ASSISTments2009}; (b) \emph{Statics2011}, where the sizes of nodes are proportional to their degrees and the colors reflect the clusters they belong to (best in colors).}
\label{fig:learned_graphs}\vspace*{-0.3cm}
\end{figure}

\begin{comment}
\begin{table}[hbt!]
  \caption{Comparing key design aspects across state-of-the-art KT models.}
  \label{tbl:aspect}
   \resizebox{\columnwidth}{!}{
  \begin{tabular}{lccc}
    \toprule
    Model&Memory (Dynamics Modelling) &Graph Representation &Forget Modelling\\
    \midrule
    BKT~\cite{Corbett1994}&Binary variable&$\times$& $\times$\\
    DKT~\cite{DKT2015}&RNN&$\times$&$\times$\\
    DKVMN~\cite{DKVMN17}& key-value memory& $\times$&$\times$\\
    DKT+forget~\cite{DKT_Forget}&RNN&$\times$&Over question space\\
    GKT~\cite{GKT}&GNN&\checkmark&$\times$\\
    SAKT~\cite{SAKT}&Attention mechanism&$\times$&$\times$ \\
    DGMN (Ours)&  key-value memory&\checkmark&Over concept space\\
  \bottomrule
\end{tabular}
}
\end{table}
\end{comment}

%\subsection{Memory augmented neural networks (MANNs)}
\medskip 
\noindent\textbf{Memory Augmented Neural Networks.~}In knowledge tracing, Memory-augmented Neural Networks (MANNs) are usually used to track temporal dynamics in a question answering sequence based on an external memory structure. This methodology is aligned with what one can find in modern computing architectures that involve processing and memory components. Graves et al.~\cite{NTM14} proposed a novel technique to attention the read and write processes on MANNs for long sequences of arbitrary information, which outperformed conventional recurrent neural models without memory. Recently, MANNs have been successfully applied to various domains such as reinforcement learning~\cite{pritzel2017neural,santoro16}, computer vision~\cite{vinyals2016matching,ConvMann19}, and time-series modelling~\cite{HTM20,TransMann19}.

\medskip 
\noindent\textbf{Graph Neural Networks.~}
Graphs representations can effectively capture the characteristics of individuals (as nodes) and their relationships (as edges) in various kinds of problems. Examples for this include social networks, protein structures, and supply chains. Scarselli et al.~\cite{GNN08} proposed a graph neural network model that can distill information from an input graph to perform supervised learning tasks. Kipf and Welling~\cite{kipf_GCN_17} proposed the graph convolutional neural networks (GCNs) model based on the concept of graph convolution. Inspired by convolutional neural networks (CNNs)~\cite{CNN_98}, GCN aggregates information from neighbour nodes into node embedding vectors, which is similar to how CNNs extract information from neighbour pixels in images. GCNs have been successfully applied in various domains such as image processing~\cite{defferrard2016convolutional}, text mining~\cite{HamiltonYL17}, and molecular modelling~\cite{GCN15}.

\medskip 
\noindent\textbf{Modelling Forgetting.~}
Cognitive studies on students learning behaviours show that forgetting is a counter aspect to knowledge gaining over-time~\cite{Cog_forget_11,Cog_13}. According to the \emph{forgetting curve theory}~\cite{ebbinghaus2013}, students have a  declining memory curve over time that can affect their knowledge proficiency unless re-practice is performed. Thus, a group of KT methods considered forgetting behaviour during the modelling process. Nagatani et al.~\cite{DKT_Forget} extended the DKT model~\cite{DKT2015} by adding sequence-related forgetting features, which showed to enhance the original model's performance. Huang et al.~\cite{Zhenya_forget_20} proposed a probabilistic matrix factorization model that considers practice priors for questions to model forgetting. Our proposed model differs from these methods by modelling forgetting behaviours over a concept space instead of a question space for achieving more accurate prediction. Wang et al.~\cite{TCEKT21} proposed a method that considers forgetting using localized temporal processes. It decomposes a sequence into local dynamic processes using the \emph{Hawkes} model and biases its predictions accordingly. % \q{as showed by example in Figure~\ref{fig:forgetting}. (is it a real example from your experiments?)}\g{This is a demo example for showing the idea on calculating forgetting features using the latent concept space, we give a real example from ASSISTments2009 dataset in Figure3}

\medskip 
\noindent\textbf{Knowledge Tracing Methods.~}
Over the last two decades, a good number of studies have attempted to address different aspects of the KT problem, in order to unlock the potential of relevant applications, such as: intelligent learning systems, curriculum learning, and learning material recommendation. These attempts can be generally categorized into three main categories: 1) Bayesian methods, 2) deep learning methods, and 3) graph-based methods. 

Bayesian inference methods~\cite{Corbett1994,Baker2008,Pardos_2011,Yudelson_2013} use state-space models~\cite{SSM_94} such as hidden Markov models (HMMs) to trace a student's knowledge state. As they apply Bayesian inference for model's parameters estimation, a simple knowledge state representation is assumed such as a binary random variable (i.e., know or do not know) to make the inference computationally tractable. While these methods are easily interpretable, their oversimplified assumptions limits the ability to track complex knowledge state dynamics.

Deep learning methods overcome this limitation by adopting more powerful representations for knowledge states. Piech et al.~\cite{DKT2015} introduced the DKT model that uses recurrent neural networks to model dynamics of knowledge states in question answering sequences. While DKT significantly outperformed Bayesian methods, it assumed a single latent concept to exist in any KT task (i.e., represents all concepts holistically). Recent methods~\cite{DKVMN17,SKVMN,AKT20} used MANNs with attention techniques to facilitate modelling of a knowledge state over multiple latent concepts.  

Despite the potential of deep learning KT methods, they usually do not have an explicit representation for relationships among latent concepts involved in the KT task. Graph-based methods~\cite{GIKT20,GKT19,HGKT_20} have been proposed, which use graph representations for latent concepts and their relationships to predict the probability of correctly answering a question. However, they often assume a predefined graph (i.e., latent concepts and their relationships), which is often difficult to obtain in real-world KT applications. Our proposed DGMN model overcomes this limitation by leveraging the information in its memory structures to dynamically construct a graph representation that captures relationships among latent concepts. Thus, DGMN can better predict the probability of correctly answering a question based on knowledge state dynamics and information from memory and graph representations.

%% file: Conclusion.tex
\section{Conclusion}
\label{sec:conclusion}
In this paper we introduced a novel model to tackle the knowledge tracing problem. This proposed model leverages knowledge state dynamics from external memory structures to construct a graph for latent concepts and their relationships, while considering student's forgetting behaviours. Thus, it can better predict student learning performance based on both knowledge temporal dynamics and relationship across latent concepts. We compared our proposed model with the state-of-the-art KT models on four well-established datasets. The results showed that our model outperforms all other models on all datasets. For the future work, we plan to investigate the potential of latent concept graphs for curriculum learning and recommending exercises to students.

\begin{comment}
\begin{table*}
  \caption{Ablation study and AUC.}
  \label{tbl:combined}
   \begin{adjustbox}{max width=\textwidth}
  \begin{tabular}{l|cccc|cccc}
    \toprule
     \multirow{2}{*}{Model}& \multicolumn{4}{c|}{Included Module}& \multicolumn{4}{c}{Dataset}\\\cline{2-9}
     &Forget&Graph&Rank&KV Memory&ASSISTments2009&Statics2011&Synthetic-5&Kddcup2010\\
      \midrule
      GKT&$\times$&\checkmark&$\times$&$\times$&$72.3\pm0.02$&$73.4\pm0.03$&$74.2\pm0.01$&$76.9\pm0.01$\\
      DKT+forget&\checkmark&$\times$&$\times$&$\times$&$73.2\pm0.02$&$74.5\pm0.01$&$75.1\pm0.03$&$79.0\pm0.01$\\
    DKVMN&$\times$&$\times$&$\times$&\checkmark&$81.6\pm0.03$&$82.8\pm0.02$&$82.7\pm0.01$&$79.8\pm0.02$\\
    SAKT&$\times$&$\times$&$\times$&$\times$&$84.8\pm0.03$&$85.3\pm0.01$&$83.2\pm0.02$&$81.7\pm0.01$\\\hline
    DGMN (ours)&\checkmark&\checkmark&\checkmark&\checkmark&$\mathbf{86.1\pm0.01}$&$\mathbf{86.4\pm0.02}$&$\mathbf{85.9\pm0.03}$&$\mathbf{83.4\pm0.01}$\\\hline
   DGMN-NoForget &$\times$&\checkmark&$\times$&\checkmark&$82.8\pm0.01$&$83.9\pm0.03$& $83.8\pm0.04$& $80.7\pm0.03$\\
   DGMN-NoGraph&\checkmark&$\times$&$\times$&\checkmark&$85.2\pm0.02$&$85.5\pm0.01$&$84.7\pm0.02$&$82.4\pm0.01$\\
   DGMN-NoRank&\checkmark&\checkmark&$\times$&\checkmark&$85.7\pm0.02$&$85.9\pm0.01$&$85.1\pm0.02$&$82.7\pm0.03$\\
   DGMN-Basic&$\times$&$\times$&$\times$&\checkmark&$81.9\pm0.01$&$82.9\pm0.03$&$82.9\pm0.01$&$80.0\pm0.02$\\
   \bottomrule
\end{tabular}
\end{adjustbox}
\end{table*}

\end{comment}